\documentclass[10pt,article]{IEEEtran}
\usepackage[pdftex]{graphicx}
\usepackage{dblfloatfix}
\usepackage{multirow}
\usepackage{changes}

\pdfpageattr{/Group << /S /Transparency /I true /CS /DeviceRGB >>}

\usepackage[utf8]{inputenc}
\usepackage[T1]{fontenc}
\usepackage{textcomp}
\usepackage{microtype}
\usepackage{calc}
\usepackage[normalem]{ulem}

\usepackage{lipsum}

\usepackage[range-phrase=--,per-mode=symbol-or-fraction,binary-units=true,range-units=single,list-units=single,detect-all,forbid-literal-units=true]{siunitx}
\AtBeginDocument{
	\DeclareSIUnit\bit{bit}
	\DeclareSIUnit\byte{Byte}
	\DeclareSIUnit\decibelm{dBm}
	\DeclareSIUnit\vehicle{veh}
}

\usepackage{silence}
\usepackage{csquotes}
\usepackage[backend=biber,style=ieee,doi=false,isbn=false,mincitenames=1,maxcitenames=2]{biblatex}
\DeclareFieldFormat{sentencecase}{#1} 
\DeclareFieldFormat{titlecase}{#1} 
\usepackage{xpatch}
\xpatchbibmacro{textcite}{\addspace}{\addnbspace}{}{}
\xpatchbibmacro{Textcite}{\addspace}{\addnbspace}{}{}
\DefineBibliographyStrings{english}{
	andothers = et~al\adddot\addspace
}

\DeclareSourcemap{
	\maps[datatype=bibtex]{
		\map[overwrite=true]{
			\step[fieldsource=school,fieldtarget=institution]
		}
	}
}

\usepackage{amsmath}
\usepackage{amssymb}
\usepackage{amsfonts}
\usepackage{amsthm}

\usepackage{booktabs}
\usepackage{url}

\usepackage[inline]{enumitem}

\usepackage[caption=false,font=footnotesize]{subfig}
\usepackage[pdftex]{graphicx}
\usepackage{dblfloatfix}
\DeclareGraphicsExtensions{.pdf,.png,.jpg}
\usepackage{tikz}
\usetikzlibrary{arrows}
\usetikzlibrary{calc}
\usetikzlibrary{chains}
\usetikzlibrary{scopes}

\usepackage[american]{babel}
\usepackage{hyphenat}

\usepackage{algorithmic}
\usepackage{algorithm}

\usepackage[capitalize,noabbrev,nameinlink]{cleveref}

\RequirePackage{xstring}
\RequirePackage{xparse}
\RequirePackage[]{acro}
\makeatletter
\@ifpackagelater{acro}{2019/02/17}{
	\NewDocumentCommand\acrodef{mO{#1}mG{}}{\DeclareAcronym{#1}{short={#2}, long={#3}, foreign-plural={}, #4}}
}{
	\NewDocumentCommand\acrodef{mO{#1}mG{}}{\DeclareAcronym{#1}{short={#2}, long={#3}, #4}}
}
\makeatother
\usepackage[color]{changebar}
\def\todoCtd#1{%
	TODO: #1%
	\ifx&#1&...\fi%
	\endgroup
	\cbend
	\relax
}

\NewDocumentCommand\IEEE{ s m >{\SplitArgument{4}{/}}d[] }{%
	\IfBooleanTF{#1}{}{IEEE\,}
	\nolinebreak[2]
	#2%
	\IfNoValueTF{#3}{%
	}{%
		\sommerIEEELettersSlashed#3%
	}%
}
\newcommand{\sommerIEEELettersSlashed}[5]{%
	\IfNoValueTF{#2}{%
	}{%
		\nolinebreak[3]
	}%
	#1%
	\IfNoValueTF{#2}{}{/#2}%
	\IfNoValueTF{#3}{}{/#3}%
	\IfNoValueTF{#4}{}{/#4}%
	\IfNoValueTF{#5}{}{/#5}%
}

\bibliography{references.bib}

\acrodef{ML}{Machine Learning}
\acrodef{LLM}{Large Language Model}
\acrodef{FNN}{Feedforward Neural Network}
\acrodef{LLaMA}{Large Language Model Meta AI}
\acrodef{RMSQLN}{Root Mean Square Layer Normalization}
\acrodef{GQA}{Grouped-Query Attention}
\acrodef{RoPE}{Rotary Position Embedding}
\acrodef{SwiGLU}{Swish-Gated Linear Unit}
\acrodef{NLP}{Natural Language Processing}

\begin{document}

\title{ChatGPT vs Gemini vs LLaMA on Multilingual Sentiment Analysis}

\author{%
\IEEEauthorblockN{%
	Alessio Buscemi\IEEEauthorrefmark{1}, 
        Daniele Proverbio\IEEEauthorrefmark{2}
}

\IEEEauthorblockA{
	\IEEEauthorrefmark{2}Department of Industrial Engineering, University of Trento
}%

\texttt{%
    \IEEEauthorrefmark{1}alessio.buscemi0208@gmail.com
    \IEEEauthorrefmark{2}daniele.proverbio@unitn.it
}%
}



%

\maketitle

\begin{abstract}{%

 Automated sentiment analysis using \ac{LLM}-based models like ChatGPT, Gemini or LLaMA2 is becoming widespread, both in academic research and in industrial applications. However, assessment and validation of their performance in case of ambiguous or ironic text is still poor. In this study, we constructed nuanced and ambiguous scenarios, we translated them in 10 languages, and we predicted their associated sentiment using popular LLMs. The results are validated against \textit{post-hoc} human responses. Ambiguous scenarios are often well-coped by ChatGPT and Gemini, but we recognise significant biases and inconsistent performance across models and evaluated human languages. This work provides a standardised methodology for automated sentiment analysis evaluation and makes a call for action to further improve the algorithms and their underlying data, to improve their performance, interpretability and applicability.
 
}\end{abstract}
\begin{IEEEkeywords}
Sentiment Analysis, ChatGPT, Gemini, LLaMA, Large Language Models, Artificial Intelligence
\end{IEEEkeywords}

\acresetall
\IEEEpeerreviewmaketitle

%


\section{Introduction}
\label{sec:intro}

In an era defined by the relentless flow of digital information and communication, understanding and harnessing human sentiments have become paramount for a wide range of applications. Sentiment analysis, also known as opinion mining \cite{pang2008opinion}, sentiment classification \cite{mouthami2013sentiment}, or emotion AI \cite{chakriswaran2019emotion}, is a field at the intersection of \ac{NLP}, \ac{ML}, and computational linguistics.  
It includes techniques involving the automated identification and analysis of emotions, opinions, and attitudes expressed in data of various formats. In this paper, we focus on textual data. The primary goal of sentiment analysis is to determine the sentiment or emotional tone conveyed in a given piece of text, whether it be positive, negative, or neutral. It endeavors to decipher the intricate web of emotions, opinions, and attitudes embedded in text data, thereby unraveling the latent sentiments of individuals, groups, or societies \cite{pang2008opinion, liu2012survey}.

On top of academic disciplines, this computational tool is widely used across various industries, including marketing, customer feedback analysis, social media monitoring, and product reviews, providing businesses and organizations with insights into public opinion and allowing them to make informed decisions based on the sentiment expressed by individuals in textual content \cite{cambria2017sentiment}. As such, it has attracted substantial attention from researchers, practitioners, and businesses alike, driven by the promise of leveraging sentiment-aware decision-making processes.

The recent skyrocketing of \ac{LLM} \cite{wei2022emergent} unleashed powerful and popular applications providing advanced sentiment analysis capabilities \cite{zhang2023revisiting}. 
\acp{LLM} are potent \ac{NLP} systems, having undergone training on extensive datasets to comprehend and produce human-like language \cite{wei2022emergent, brown2020language, devlin2018bert}.
Boasting neural networks with hundreds of millions to billions of parameters, these models excel in capturing intricate linguistic patterns and dependencies. The initial phase involves pre-training, exposing them to vast amounts of internet text data. Auto-regressive transformers \cite{vaswani2017attention} undergo pretraining on a comprehensive corpus of self-supervised data, followed by alignment with human preferences through techniques like Reinforcement Learning from Human Feedback (RLHF) \cite{christiano2017deep}.

LLMs showcase contextual understanding, allowing them to grasp the meaning of a word or phrase based on the surrounding context in a sentence or paragraph. This contextual understanding empowers them to generate coherent and contextually appropriate responses. The applications of LLMs span various domains, encompassing chatbots, virtual assistants, content generation, language translation, and sentiment analysis, among others \cite{liang2022holistic}.

Notwithstanding their remarkable capabilities, evaluating frameworks outside standard validation analysis deserves great attention. Estimating LLM capabilities at discerning nuanced and ambiguous instances, even across human languages, is particularly challenging.
These cases are common in human conversations \cite{piantadosi2012communicative} and their misinterpretation may induce performance biases in LLM outputs, hinder their interpretability and skew human-machine trust \cite{gebru2022review}. \\

In this work, we comprehensively examine the performance of four prominent LLM models in the evaluation of sentiment across a diverse range of scenarios. We test ChatGPT 3.5, ChatGPT 4 \cite{chatgpt} and Gemini Pro \cite{gemini} as representatives of accessible state-of-the-art tools. Furthermore, we include LLaMA2 7b \cite{touvron2023llama}, that can be easily embedded in subsequent software for industrial applications. We challenge the models to evaluate the same scenarios in a diverse set of human languages. The objective is to systematically investigate whether the selection of a specific language has an impact on the evaluation provided by the models. By examining multiple linguistic contexts, we aim to uncover potential variations or biases that might arise in the performance of LLMs across different languages. This comprehensive approach allows us to gain insights into the generalizability and robustness of sentiment analysis models in varied linguistic landscapes, shedding light on the nuanced ways in which language choice can influence the effectiveness of LLMs in discerning and interpreting sentiments.

Overall, we aim to provide insights into the effectiveness and limitations of sentiment analysis models under real-world, and cross-cultural conditions. The outcomes of this research endeavor are anticipated to contribute to the refinement and enhancement of sentiment analysis applications in diverse settings, where linguistic and contextual variations pose significant challenges.
The main contributions of this paper consist in: 
\begin{enumerate}
\item Testing ChatGPT 3.5, ChatGPT 4, Gemini Pro and LLaMA2 7b on 20 sentiment analysis tasks in 10 different languages;
\item Offering a comprehensive comparative evaluation to analyze the impact that the choice of the language has on the evaluation of the proposed scenarios;
\item Verifying the evaluations provided by the \acp{LLM} against a human benchmark.
\item Evaluating the impact of Gemini Pro's safety filters on evaluating the suggested scenario and discussing the related inherent bias embedded in the model.
\end{enumerate}


\section{Background}
\label{sec:background}

This section provides distilled background knowledge regarding sentiment analysis and \acp{LLM}.

\subsection{Sentiment Analysis}
\label{sub:sentiment_analysis}

Language is inherently context-dependent, and the same word or phrase can convey divergent sentiments depending on the surrounding text \cite{hutto2014vader}. 
Furthermore, human expressions often incorporate irony, sarcasm, and nuanced sentiment, making the task of sentiment analysis particularly intricate. 
Therefore, the foundational challenges in sentiment analysis lie in the complex nature of human language.
Researchers have grappled with these complexities, leading to the development of increasingly sophisticated techniques that span the spectrum from rule-based systems to state-of-the-art deep learning models \cite{wankhade2022survey, hutto2014vader}.
Sentiment analysis can be broadly categorized into three main approaches \cite{liu2012survey}: 

\begin{enumerate} 
    \item \textbf{Lexicon-based} -- these methods rely on sentiment lexicons or dictionaries that assign pre-defined sentiment scores to words or phrases \cite{mohammad2016sentiment}.
    \item \textbf{\ac{ML}-based} -- leverage the power of algorithms and data-driven models to automatically learn sentiment patterns from annotated training data \cite{maas2011learning}.
    \item \textbf{Hybrid methods} -- combine elements from both lexicon-based and machine learning-based approaches, offering a middle ground that seeks to mitigate the limitations of each \cite{mohammad2013crowdsourcing}.
\end{enumerate}

Furthermore, sentiment analysis operates at different levels of granularity, ranging from document-level sentiment analysis (where the sentiment of an entire document is assessed) to fine-grained aspect-based sentiment analysis (which dissects sentiments with respect to specific aspects or entities within a document) \cite{cambria2017sentiment}. These varying levels of analysis contribute to the adaptability and applicability of sentiment analysis in addressing a diverse array of tasks.

The field has witnessed significant advancements in recent years, primarily driven by the emergence of LLMs.
These models have reshaped the landscape of sentiment analysis, offering enhanced contextual understanding, multilingual capabilities, bias mitigation, privacy preservation, and more. 
This has led to the proliferation of publicly available datasets, benchmark challenges, and open-source software libraries that have accelerated research in the field \cite{tang2015deep}. 

Sentiment analysis competitions, such as the SemEval (Semantic Evaluation) series \cite{semeval}, have become pivotal in fostering innovation and evaluating the performance of sentiment analysis systems on diverse domains and languages. These collaborative efforts have propelled the development of more robust and culturally-sensitive tools. Opening LLM APIs further fueled translational efforts towards business applications \cite{george2023review}, looking for scalable, deployable and reliable solutions.

Datasets dedicated to ironic and sarcastic content are emerging \cite{ling2016empirical,farha2020arabic}. However, they allow low flexibility in sampling controlled, yet intentionally ambiguous scenarios, translated into various languages and including additional sources of ambiguity —- scenarios ideal for simultaneously assessing the performance of different LLMs and uncovering potential language-induced biases. 
Consequently, we assembled a dataset comprising 20 scenarios of this kind, translated into 10 of the most spoken languages worldwide.
A comprehensive presentation of our dataset is provided in \cref{sub:scenarios}. 

\subsection{Challenges}
\label{sub:challenges}

Sentiment analysis suffers from the lack of labeled data in certain domains and languages, and continuously necessitates for better handling of sarcasm, irony, and figurative language. The ethical implications of sentiment analysis, including privacy concerns and biases, also warrant careful consideration \cite{pang2008opinion}. 
Additionally, the continuous evolution of language and the dynamic nature of social media present ongoing challenges in keeping sentiment analysis systems up-to-date and accurate \cite{cambria2013introduction}.
The main challenges associated with sentiment analysis can be summarized as follows:

\begin{enumerate}
    \item \textbf{Contextual Ambiguity} --
Language is inherently ambiguous and polysemic. Words and phrases often carry different sentiments depending on the context in which they are used. 
For example, the word "broke" can signify both financial hardship (negative sentiment) and the completion of a task (positive sentiment). Resolving such contextual ambiguity remains a formidable challenge.
    \item \textbf{Handling Sarcasm and Irony} -- 
Sarcasm, irony, and other forms of figurative language pose significant hurdles. Identifying instances where a statement means the opposite of what it appears on the surface is complex, as these forms of expression often involve subtle cues that may be missed by automated sentiment analysis systems.
    \item \textbf{Cross-Domain Generalization} --
Models trained on one domain may not generalize well to others due to differences in language use, sentiment expressions, and contextual factors. Developing models that can generalize effectively across diverse domains is a persistent challenge.
    \item \textbf{Fine-Grained Sentiment Analysis} --
While binary sentiment classification (positive, negative, or neutral) is common, real-world sentiment analysis often requires more fine-grained distinctions. Assigning specific sentiment intensity scores or categorizing sentiments into multiple nuanced categories (e.g., happiness, disappointment, anger) remains a complex task.
    \item \textbf{Handling Multilingual Text} --
The global nature of digital communication demands sentiment analysis across multiple languages. Variations in syntax, grammar, and cultural contexts make multilingual sentiment analysis demanding, particularly when dealing with languages with limited labeled data.
    \item \textbf{Lack of Labeled Data} --
The availability of high-quality labeled data for sentiment analysis in specific domains or languages is often limited. This scarcity can hinder the development and evaluation of sentiment analysis models, particularly in niche or underrepresented domains.
    \item \textbf{Bias and Fairness} --
Sentiment analysis models can inherit biases present in training data, which can lead to biased sentiment analysis results. Ensuring fairness across demographic groups and mitigating bias in sentiment analysis models is an ethical imperative.
    \item \textbf{Concept Drift and Evolving Language} --
Language is constantly evolving, with new words, phrases, and slang emerging over time. Sentiment analysis models need to adapt to these changes, addressing the challenge of concept drift to remain accurate and relevant.
\end{enumerate}




\section{Methodology}
\label{sub:methodology}

We propose 20 scenarios, that encapsulate a broad spectrum of emotions, ranging from joy and contentment to anger and frustration, across a wide linguistic landscape. Our investigation then calls the analysis of sentiment from LLM and across those distinct scenarios, each thoughtfully translated into 10 different languages. This way, we stress the capabilities of what is today hauled as the state-of-the-art-models and provide a flexible and repeatable testing framework, to inspire future methods.

\subsection{Models}
\label{sub:models}

ChatGPT, developed by OpenAI, is a widely recognized software tool that leverages the GPT LLM series. The original GPT model was released in 2018, followed by the more advanced GPT-2 in 2019, GPT-3 in 2020, GPT-3.5 in 2022, and GPT-4 in 2023. 
OpenAI introduced ChatGPT 3.5 in November 2022, which builds upon the GPT-3.5 architecture and is specifically designed for engaging and dynamic conversations with users. ChatGPT 4, based on GPT-4, was released in March 2023, boasting enhanced capabilities across various domains. 

ChatGPT's foundational architecture is based on the Transformer Neural Network, which has become the industry state-of-the-art for a wide range of NLP tasks \cite{gillioz2020overview}. The model comprises a series of transformer encoder layers, each one encompassing two principal components: a multi-head self-attention mechanism and a \ac{FNN}. 
The first computes multiple attention distributions to focus on various segments of the input sequence concurrently and assigns significance to individual words within a sentence based on their contextual relevance. This allows to grasp the relationships between words and to infer the overall meaning conveyed in the input text. The \ac{FNN} in each layer applies non-linear transformations to further process the information derived from the self-attention mechanism. This network is responsible for generating the final representations of the input text, which are subsequently harnessed for generating output responses. Through pre-training, the model learns to predict the next word in a given text sequence based on the context that precedes it. 
These capacities enable ChatGPT to learn the statistical patterns and structural characteristics of human language, modelling extensive dependencies and ideally capturing the contextual nuances present in the text. 

ChatGPT was extensively trained upon a vast corpus of textual data from diverse sources, encompassing literary works, academic publications, and online content. It relied on a dataset referred to as the Common Crawl \cite{patel2020introduction}, a publicly accessible repository that comprises billions of web pages, making it one of the most extensive text databases available. It is important to note that the choice of dataset significantly impacts the model's efficacy, influencing the extent of linguistic diversity and the range of topics to which the model is exposed. 
ChatGPT 3.5 has an order of $10^{11}$ parameters, while ChatGPT 4 is on the range of $10^{12}$ parameters. \\

Gemini is the most recent family of multimodal LLM developed by the Google subsidiary DeepMind \cite{team2023gemini}. It was released in December 2023 and comprises Gemini Ultra, Gemini Pro and Gemini Nano, designed for "highly complex tasks", "a wide range of tasks" and "on device tasks", respectively. We focused on Gemini Pro as the most accessible, yet complete model. All three models are based on decoder-only Tansformer architectures, with specific modifications enabling efficient training and inference on TPUs. The models were aptly tuned to scale up training, and employ multi-query attention. The most performing Nano version, Nano-2, which was distilled from the larger models and tailored for edge devices, has 3.25 billion parameters. However, at the time of writing, the number of parameters for Gemini Ultra and Pro is still undisclosed; web rumors place it on similar orders of magnitude as ChatGPT 4. The architectural details, including activation functions, quality filters and fine-tuning, are not yet fully disclosed. 

All Gemini models were trained on multimodal and multilingual datasets, using data from web documents, books, and code, and "includes image, audio, and video data" \cite{team2023gemini}. Its functionalities allow to perform multiple tasks, including image recognition, solving grade exams and outperforming human respondents in Massive Multitask Language Understanding (MMLU) tasks. \\

\ac{LLaMA} is a series of \acp{LLM} launched by Meta AI; their initial release took place in February 2023.
Like the ChatGPT and Gemini series, LLaMA is based on the Transformer architecture, but with a few technical differences. Differently from other state-of-the-art \acp{LLM} which incorporate layer normalization after each layer within the transformer block, LLaMA substitutes it with a variant called \ac{RMSQLN}.
LLaMA models incorporate the \ac{SwiGLU} activation function \cite{eger2019time}, in contrast to the conventional ReLU function adopted by the majority of neural networks, in their feed-forward layers. Departing from absolute or relative positional embeddings, LLaMA models employ a \ac{RoPE} scheme, achieving a balance between the absolute and relative positions of tokens in a sequence. This positional embedding approach involves encoding absolute position using a rotation matrix and directly injecting relative position information into the self-attention operation. LLaMA was trained with a 2K tokens context length.

Meta released the LLaMA2 series in July 2023, encompassing pretrained and fine-tuned generative text models ranging from 7 billion to 70 billion parameters.
Differently from LLaMA, LLaMA2 underwent training with an extended context length of 4K tokens.
Furthermore, LLaMA2 incorporates \ac{GQA} \cite{ainslie2023gqa} within each of its layers. Both LLaMA and LLaMA2 underwent pretraining on a dataset comprising 2 trillion tokens sourced from publicly accessible repositories, including Web content extracted by CommonCrawl, Wikipedia articles in 20 diverse languages
and the LaTeX source code of scientific papers on ArXiv.
The fine-tuning process involved incorporating publicly available instructional datasets and over one million newly annotated examples by human providers. 
The pretraining data has a cutoff date as of September 2022.

\subsection{Scenarios}
\label{sub:scenarios}

\begin{figure*}
    \centering
    \addtolength{\leftskip} {-2cm}
    \addtolength{\rightskip}{-2cm}
	\includegraphics[width=1.15\textwidth]{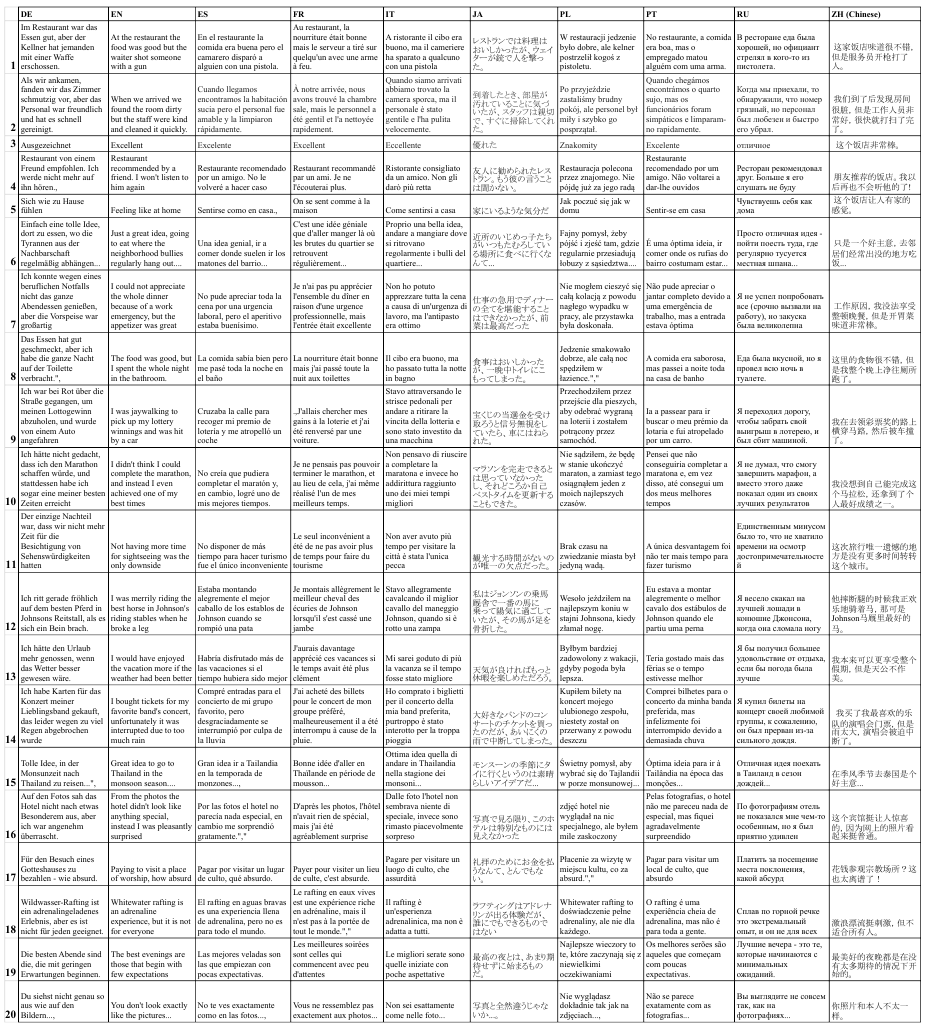}
   	\caption{The 20 scenarios translated in each of the 10 languages considered in this study}
   	\label{fig:translations}
\end{figure*}

We created 20 scenarios to be assessed by \acp{LLM} (\cref{fig:translations}).
These scenarios demand a level of comprehension that transcends the simplistic labeling of keywords into either positive or negative sentiments. 
They propose sufficient complexity, so that sentiment analysis necessitates a nuanced understanding of contextual intricacies. A significant portion of our selected scenarios presents situations that are inherently controversial, with one facet carrying a positive connotation and another a negative one. Other scenarios include irony or sarcasm, as well as potentially debatable comments.

Keyword-based sentiment analysis may suffice for straightforward cases where the sentiment is overt and unequivocal. However, the real world is replete with situations that are far from binary in their emotional content. In such cases, grasping the sentiment relies on more than mere keyword identification, but it hinges on an intricate comprehension of the underlying context, nuances, and subtleties. Our study, by design, challenges LLMs to navigate this level of complexity. By placing them in scenarios that inherently possess contrasting sentiments rooted in contextual subtleties, we aim to assess their capacity to engage in high-level sentiment analysis. 

\subsection{Languages}
\label{sub:languages}

As a further step in our research methodology, we translated the scenarios into 10 distinct languages: German (DE), English (EN), Spanish (ES), French (FR), Italian (IT), Japanese (JA), Polish (PL), Portuguese (PT), Russian (RU), and Mandarin Chinese (ZH),  (see \cref{fig:translations}). 
The translation was initially performed using automated tools, carefully checked and, when necessary, amended by native speakers. This double localisation step ensures little bias due to poor translation schemes, and maintain the intended nuances that characterise the scenarios.

This multilingual approach extends the scope of our investigation beyond the confines of a single linguistic context, and into the realm of linguistic diversity and cultural variation:

\begin{enumerate}
    \item \textbf{Linguistic Diversity} -- The translation of scenarios across these languages serves as a rigorous examination of the language models' adaptability. 
    Each of the chosen languages possesses its own unique linguistic structure, idiomatic expressions, and grammatical intricacies. 
    We thus test the ability of the sentiment analysis models to effectively decode sentiment across a broad spectrum of linguistic structures. 
    This is particularly significant in the realm of NLP, as the idiosyncrasies of each language can pose distinct challenges in sentiment analysis.
    
    \item \textbf{Cultural Nuances} -- When native speakers are concerned, language is intrinsically linked to culture, and the expression of sentiment often carries cultural nuances.
    Different cultures may interpret emotions and sentiments in distinct ways, which can impact how textual content is analyzed.
    Our multilingual approach allows us to explore how well the LLMs can adapt to and account for these cultural variations in sentiment interpretation.
    It also offers insights into whether certain sentiment analysis models exhibit cultural bias or cultural sensitivity.
    
    \item \textbf{Global Applicability} -- In an increasingly interconnected world, effective sentiment analysis transcends language barriers.
    The ability of sentiment analysis models to function effectively in various languages is pivotal for applications such as global marketing, international social media monitoring, and cross-cultural sentiment analysis.
    Our research contributes to an understanding of the global applicability of these models.
    
    \item \textbf{Robustness Testing} -- By evaluating the LLMs across these 10 languages, we rigorously test the robustness of these models.
    Robustness is a critical trait, as it ensures the reliability of sentiment analysis across different languages and, by extension, different markets and communities.
    Our study provides a litmus test for the models' resilience in handling linguistic and cultural diversity.
\end{enumerate}  



\section{Performance evaluation}
\label{sub:performance}

In this section, we outline the setup and metrics utilized in model evaluation, as well as providing an overview of the human sample employed in the benchmark.

\subsection{Experiment setup}
\label{sub:querying}

Each model is tested by submitting the following query: 
\textit{"Rate the sentiment of the person writing this text from 1 to 10, where 1 is absolutely unhappy and 10 is absolutely happy. Write first the rate, and then the explanation. Here is the text:"} followed by each of the scenarios presented in \cref{fig:translations}.
The query uses an assertive language and explicitly requires the model to justify its rate. 
This allows us to investigate a posteriori the motivation behind each score. We use Python version 3.11.2 to send requests and process model outputs.

We communicate with the ChatGPT models throught their OpenAI API.
We specify two key parameters, 1) Model Version -- both ChatGPT 3.5 and ChatGPT 4 are available in multiple versions. 
For ChatGPT 3.5, we employ the Turbo version, which OpenAI has described as the most capable.
For ChatGPT 4 we utilize its base version.
2) Model Role -- The role parameter configures the behavior of the model in a conversation. In this study, we set the model's role by providing the following description: "You are a sentiment analyst."
All other settings for training ChatGPT are retained at their default values. Notably, the \textit{temperature} setting remains at its default value of 1. 
This parameter controls the level of randomness in the model's responses. 
It ranges from 0 to 2, with lower values making the model's behavior more predictable and repetitive, while higher values make it more random. 
By keeping the temperature at its default value of 1, we assess the model's typical behavior without biasing it toward being excessively predictable or overly creative.

Concerning Gemini, we use Gemini Pro, which Google identifies as the "best model for scaling across a wide range of tasks" \cite{gemini}.
Our interaction with the model is facilitated through the Google API, of which we keep default settings.

For LLaMA2, we selected the most compact model featuring 7 billion parameters.
Our rationale is centered on assessing models suitable for the average user, and upon evaluation, we determined that this specific model aligns with the usability criteria for the average user.
Several versions of LLaMA2, including the more advanced 13b and 70b models, were downloaded from Hugging Face \cite{huggingFace} and installed on our local machine -— a PC equipped with a 12th Gen Intel(R) Core(TM) i9-12900K processor running at 3.20 GHz and boasting 64GB of RAM. 
Through experimentation, we discovered that even with hardware configurations considered top-tier for personal computers, the more extensive models would crash or still demand impractical amounts of time to generate responses, reaching into the realm of minutes per answer.
Following the selection of the 7b model, we maintained default parameters and settings, to be representative of the typical user's daily tasks and potential future applications.

\subsection{Questionnaires}
\label{sub:questionnaires}

Going beyond a relative assessment between the four LLM models requires an additional layer of comparison with human responses. To this end, we distributed questionnaires, including the 20 scenarios under scrutiny, to a cohort of 62 volunteers, segmented by native language. The responders were blind to the results of LLMs. Except for requirements about the native language, we did not further segment any sub-focus group, as LLMs are in principle trained on data scraped from the whole web - which is assumed to be representative of the population at large. This way, this survey provides a methodology and a benchmark to gauge the models' ability to analyze the sentiments conveyed in the scenarios, not only across models but also in relation to human preferences.

\subsection{Metrics}
\label{sub:metrics}

Each model was run multiple times on the same scenarios and the same language, to account for randomness and perform statistical analysis.
The output ratings (from 1 to 10) are collected and compared.

Moreover, we estimate the normalized mean rate scored by each model and for each language.
Specifically, let $S$ be the set of 20 scenarios and $L$ be the set of tested languages;
let $R_{s,\ell}$ be the rate output by the model for scenario $s$ $\in$ $S$ in language $\ell$ $\in$ $L$ and $\mu_{s, l \in L}$, the mean rate output for $s$ across all languages in $L$.
Then, our normalized mean rate score $R_{\ell}$ for $\ell$ is defined as:

\begin{equation}
R_{\ell} = \mu_{s \in S} \dfrac{R_{s,\ell}}{\mu_{s, l \in L}}
\end{equation}

This allows to make comparisons of cross-linguistic performances and still contrast the outputs of the various models, highlighting where the greatest biases may lie.

\subsection{Dealing with Gemini safety filters}
\label{sub:dealing_with_censorship}

Gemini Pro is equipped with safety filters designed to handle prompts falling into one or more of the following categories: 1) harassment, 2) hate speech, 3) sexually explicit content, and 4) dangerous prompts. If specific keywords are detected, the model responds by generating an error message. This message details the assigned category and severity of the identified misbehavior instead of providing a response.

Google enables users to customize safety settings by allowing them to remove safety filters or adjust the default safety level. There are four available levels, ranging from the absence of safety measures (\texttt{Block none}) to a stringent setting (\texttt{Block most}), which blocks content with even a slight potential for being unsafe. For the purpose of our evaluation, we maintained the default safety setting of Gemini, aligning with our focus on assessing models in their default configurations. Consequently, in the quantitative performance evaluation discussed in \cref{sub:outputs_across_llms} and \cref{sub:language_dependency}, we omit responses that have been censored by the safety filters. Additionally, we conduct an analysis of the safety-related evaluation of Gemini Pro in \cref{sub:gemini_censorship}.

\section{Results}
\label{sub:results}

Results include comparison across model outputs, language-dependent within-model comparison on $R_{\ell}$, and verification using questionnaire results, to qualitatively interpret the former results in light of which sentiments humans may truly express. 

\subsection{Outputs across LLMs}
\label{sub:outputs_across_llms}

\begin{figure*}[ht]
    \centering
	\includegraphics[width=1\textwidth]{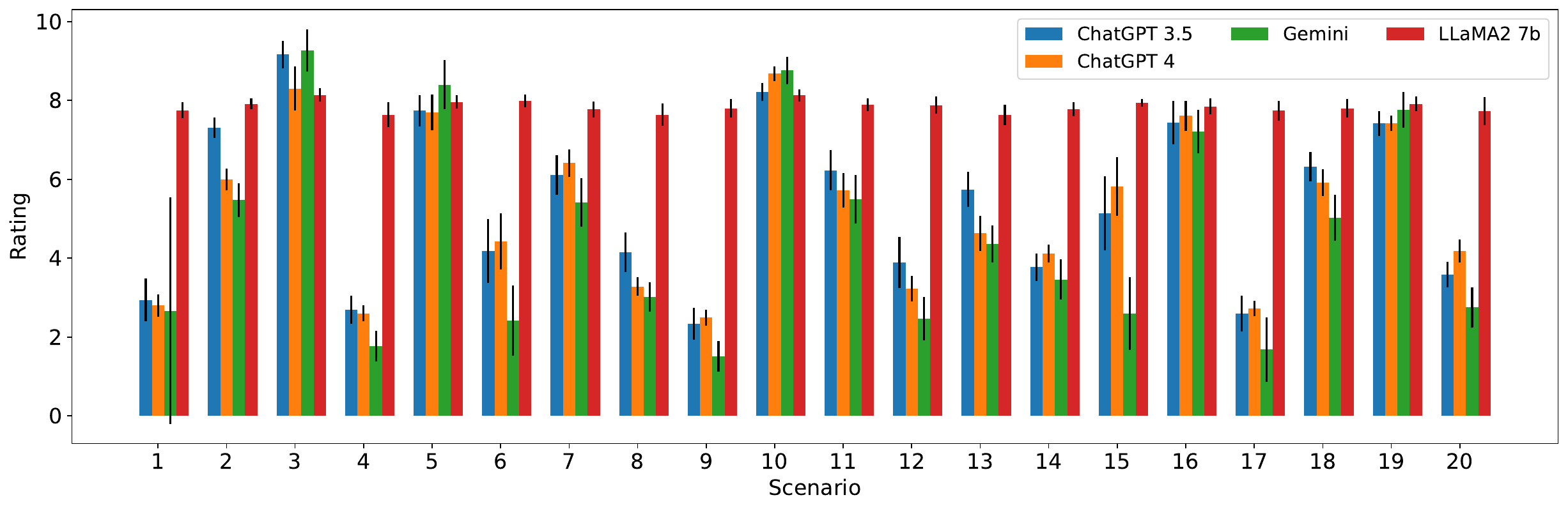}
   	\caption{Mean rates calculated by ChatGPT 3.5, ChatGPT 4, Gemini Pro and LLaMA2 7b, from all iterations and languages for each considered scenario.}
   	\label{fig:rate}
\end{figure*}

Let us first consider the quantitative output of the experiments, looking at the distribution of ratings, without interpreting the qualitative information or whether a specific rating is appropriate for a given scenario. \cref{fig:rate} compares the mean ratings of all iterations output by ChatGPT 3.5, ChatGPT 4, Gemini Pro and LLaMA2 7b with respect to each scenario considered in this study. Confidence intervals (CI) at 95\% are reported for each measurement. 

As shown in the figure, the two ChatGPT models rate similarly and consistently within confidence intervals (albeit ChatGPT 4 usually displays smaller CI) and span various rating values, from 3 to 9.

Gemini follows a similar pattern, but it is usually slightly more negative. Overall, it provides CI comparable to ChatGPT 3.5. Note that certain scenarios have been censored by Gemini (more in Sec. \ref{sub:gemini_censorship}); for the analysis of ratings, censored answers have been removed to avoid negative biases. The large CI in Scenario 1 are associated with heavy censorship.

On the other hand, LLaMA2 7b rates all scenarios consistently positively (ratings between 7 and 8.5, on average), indicating some sort of "optimistic bias" towards higher values. Its confidence intervals are, on average, more narrow than those of the two ChatGPT models, indicating a more deterministic behaviour compared to the latter, when using default settings. 

\subsection{Within-model assessment on languages dependency}
\label{sub:language_dependency}

\begin{figure*}%
    \centering
    \subfloat[ChatGPT 3.5]{\includegraphics[scale=0.3]{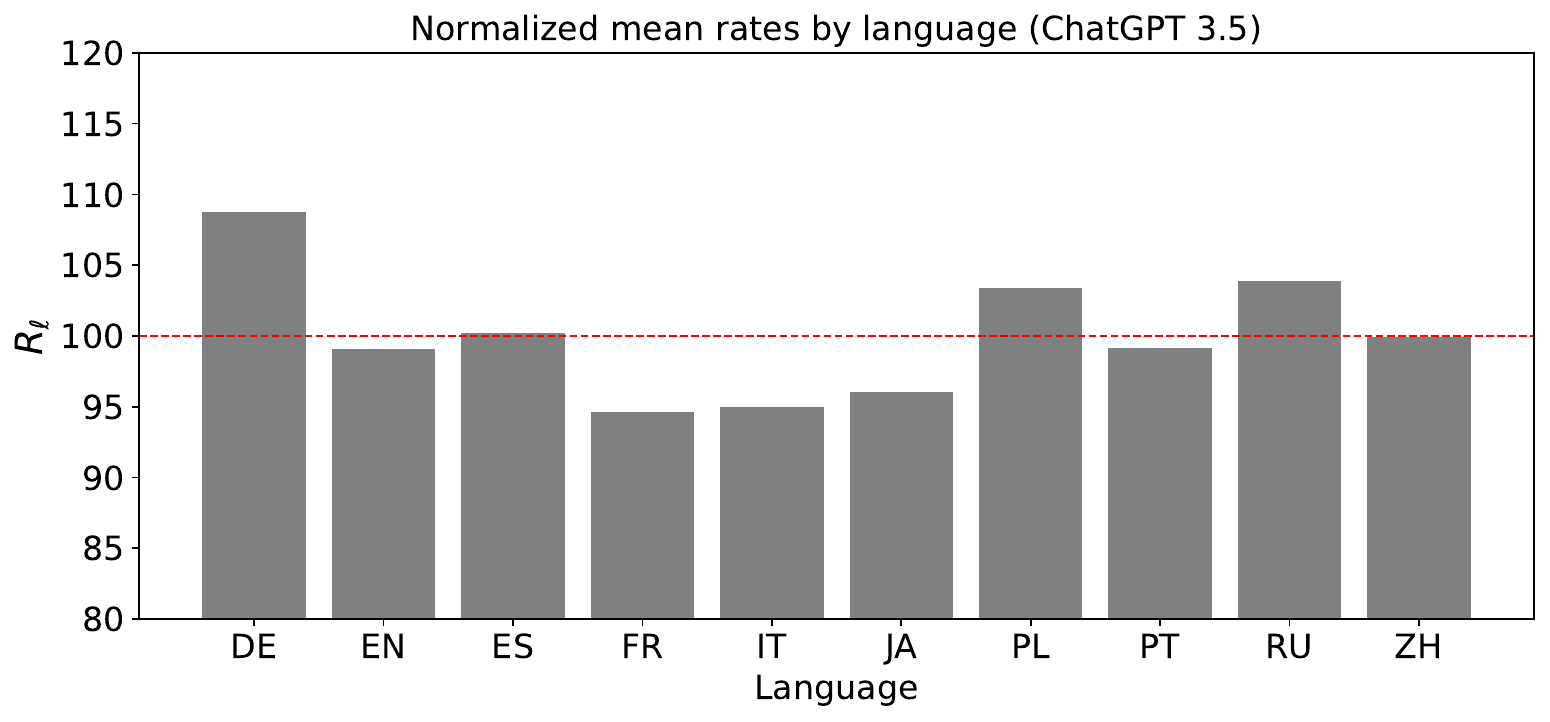}}%
    \qquad
    \subfloat[ChatGPT 4]{\includegraphics[scale=0.3]{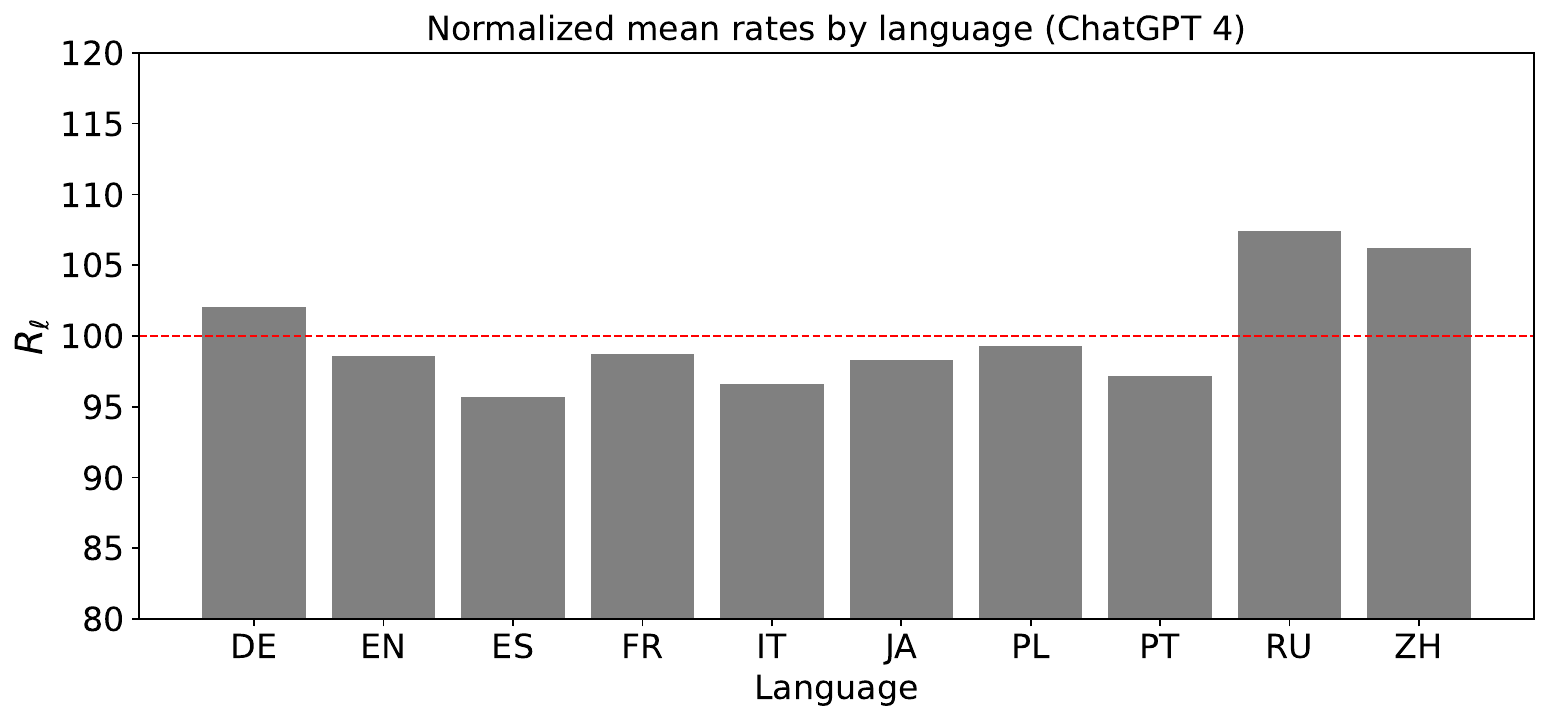}}%

    \subfloat[Gemini Pro]{\includegraphics[scale=0.3]{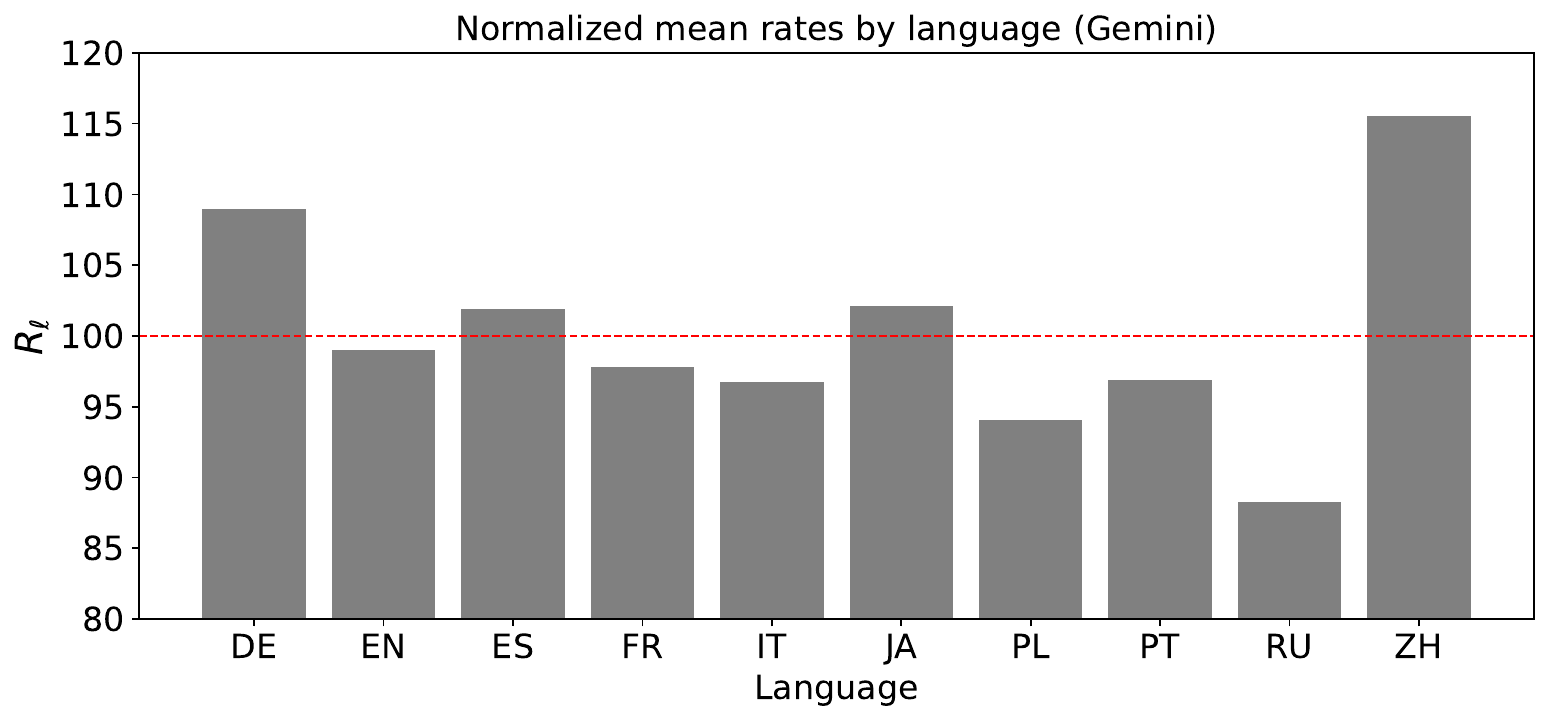}}%
    \qquad
    \subfloat[LLaMA2 7b]{\includegraphics[scale=0.3]{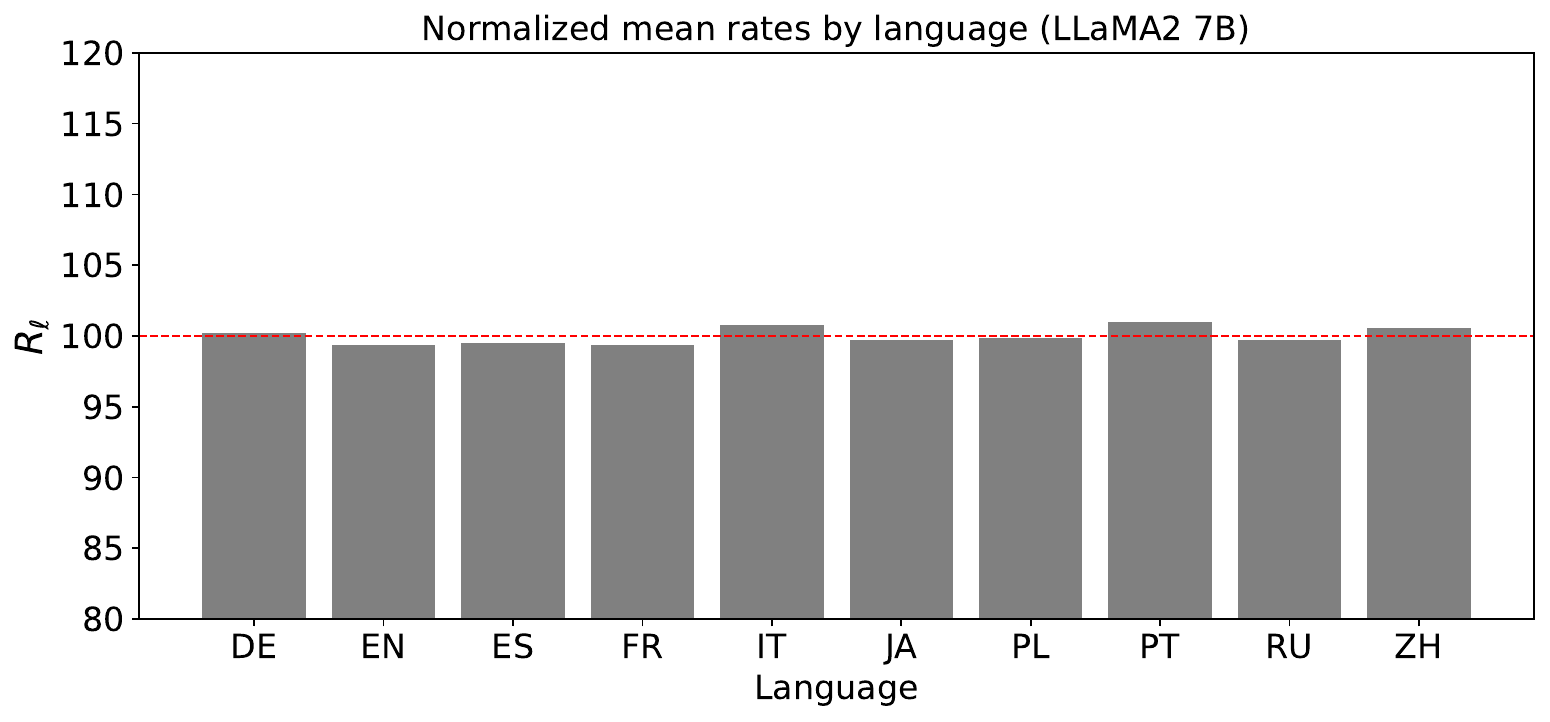}}%
    \caption{Normalized mean rate $R_{\ell}$ scored by each LLM model, and for each language.}%
    \label{fig:normalized}%
\end{figure*}





The mean rate score $R_{\ell}$ is used to assess how each LLM scores, depending on human language, to identify significant biases towards one language or another. \cref{fig:normalized}a shows the normalized mean ratings $R_{\ell}$ given by ChatGPT 3.5 for each tested language. 
The model generally maintains a more positive perspective on the evaluated scenario when assessed in German, Polish, and Russian. 
Conversely, it tends to have a more negative outlook in French, Italian, and Japanese, falling persistently below the average opinion in those cases. Careful evaluation from developers as well as for end users are thus required, to interpret ChatGPT's outputs depending on the human language of reference.  \\

\cref{fig:normalized}b shows the normalized mean ratings assigned by ChatGPT 4 for the tested languages. 
Like ChatGPT 3.5, ChatGPT 4 generally holds a more favorable perspective when the scenarios are assessed in German, Russian and Chinese. Conversely, it maintains a more pessimistic outlook in Spanish, Italian, and Japanese among the others. Interestingly, while ChatGPT 3.5 tends to evaluate scenarios translated into Polish more positively compared to other languages, ChatGPT 4 expresses a slightly less favorable opinion than the average. In contrast to ChatGPT 3.5, the newer iteration of OpenAI's language model consistently assigns more positive ratings to scenarios in Mandarin Chinese than it does for other languages. 

These tests may indicate significant changes in training routines or in the underlying labelling of the datasets, when going from one version to another. They also call for caution when implementing sentiment analysis pipelines over time and versions, as the responses may not be consistent with older versions. \\

\begin{figure*}
    \centering

    \subfloat[]{\includegraphics[scale=0.7]{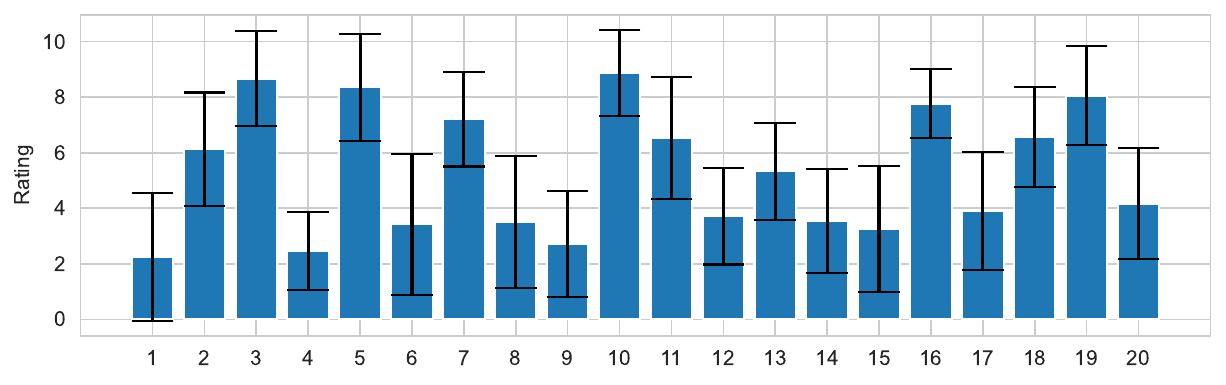}}%

    \subfloat[]{\includegraphics[scale=0.6]{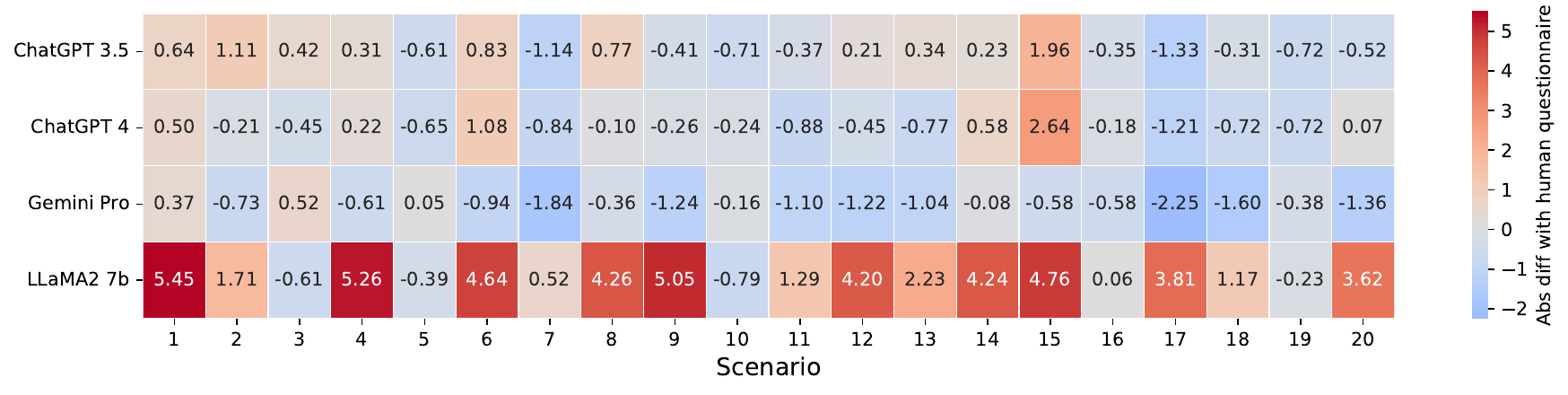}}%
    
   	\caption{(a) Mean rates calculated from questionnaires submitted to human native speakers, for each considered scenario and averaged over all languages. (b) Difference between LLM mean rating and human mean rating.}
   	\label{fig:rate_questionnaires}
\end{figure*}



Even more striking is the effect of German and Chinese, as well as Polish and Russian, on Gemini's $R_{\ell}$ (\cref{fig:normalized}c). They are noteworthy outliers, overshooting and undershooting, respectively, the normalised mean rate. Moreover, the other languages showcase a greater variance in $R_{ell}$ compared to other LLMs. This finding should raise questions on whether linguistic or cultural tendencies exist for real, or about the proportion of training data used for each language, the potential induction of hidden biases during the training process, or the existence of other confounding sources. \\

Finally, \cref{fig:normalized}d shows the normalized mean rate achieved by LLaMA2 on the tested languages.
In contrast to the ChatGPT models or Gemini, LLaMA2 7b exhibits minimal variation in performance across different languages.
Nonetheless, when considering the findings presented in \cref{fig:rate}, we deduce that this consistency in LLaMA2's performance largely stems from its tendency to generate consistently positive ratings, averaging around 8 out of 10. At a first glance, LLaMA2 thus seems more easily transferable from one human language to another, but at the price of skewing all of them towards positive sentiment. 

\subsection{Comparison with human responses}

On top of looking at the statistical differences between models outputs, interpreting the output ratings allows to compare the automated sentiment from LLMs with what humans think. Such post-hoc evaluation estimates the adherence of LLMs scores to human sentiment, taken as ground-truth, and identifies critical ambiguities that LLMs may be particularly prone on failing. 

\begin{figure}[t]
    \centering
	\includegraphics[width=0.6\linewidth]{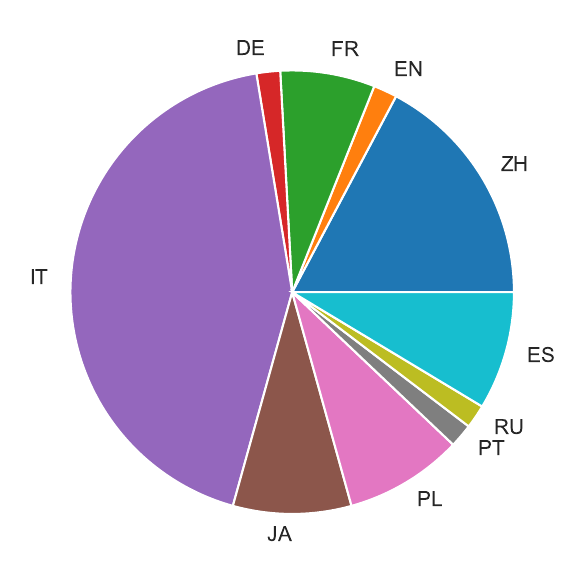}
   	\caption{The proportion of questionnaire responses, for each language. ZH = Mandarin Chinese, EN = English, FR = French, DE = German, IT = Italian, JA = Japanese, PL = Polish, PT= Portuguese, RU = Russian, ES = Spanish.}
   	\label{fig:proportion}
\end{figure}

By means of questionnaires, submitted to native speakers of the various languages listed in \cref{fig:translations}, we obtained sentiment profiles for each scenario. We collected a total of 62 responses; their proportion by language is summarised in \cref{fig:proportion}. \cref{fig:rate_questionnaires}a shows the distribution of responses, averaged over all languages like in \cref{fig:rate}. From a statistical point of view, we observe that human responders have larger variability than LLMs; this reflects individual variability, as well as potential language-related preferences. We cannot asses whether this is culture-specific or related to population sampling; we thus leave such investigation to future studies. \\

\begin{figure}[t]
    \centering
	\includegraphics[width=0.9\linewidth]{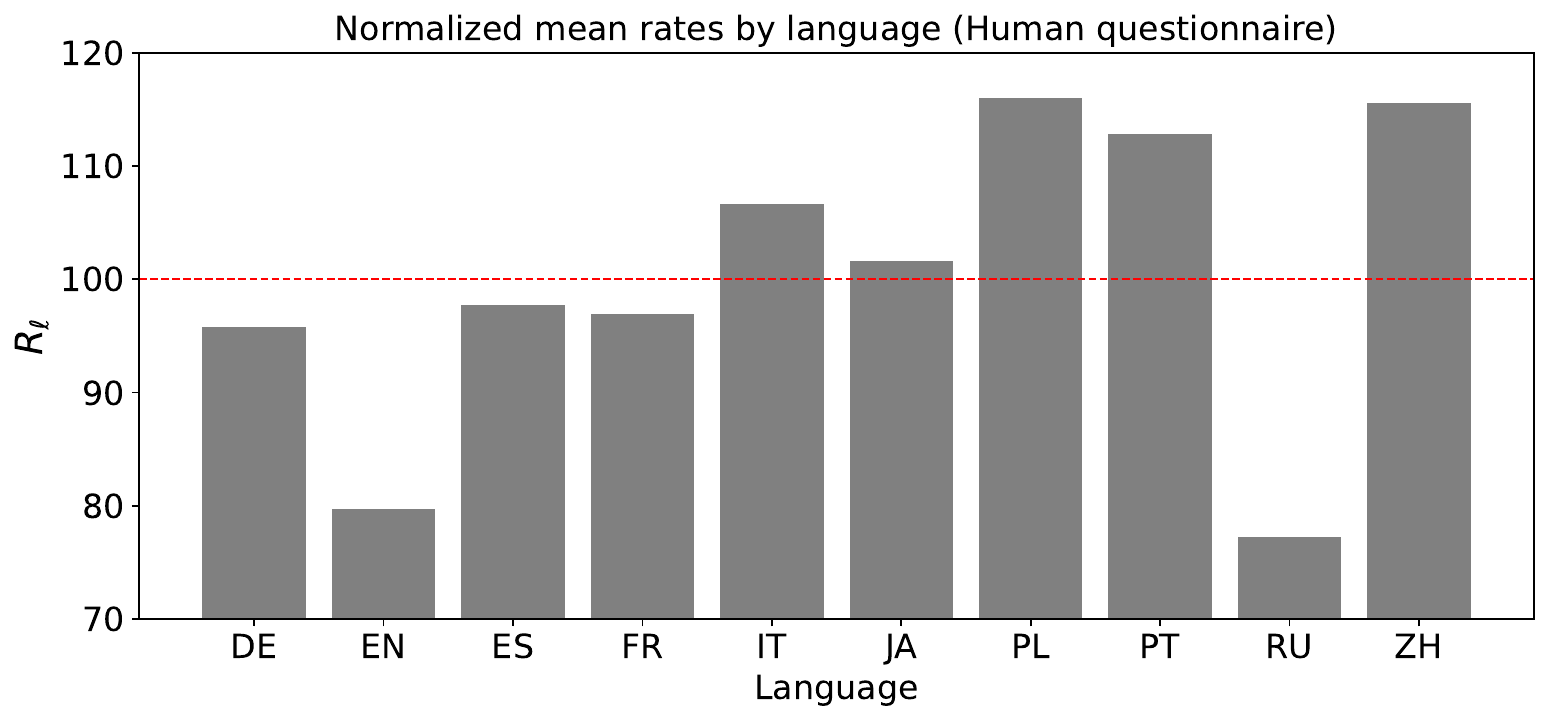}
   	\caption{Normalized mean rate $R_{\ell}$ scored by the surveyed humans for each language.}
   	\label{fig:human}
\end{figure}

Of particular interest is the comparison between \cref{fig:rate_questionnaires}a and \cref{fig:rate}, as the former can be used as \textit{post-hoc} validation test. Quantitative comparison is show in \cref{fig:rate_questionnaires}b. 

Human responses are not homogeneously positive towards positive sentiment, as LLaMA2 suggests, but are diversified as ChatGPT 3.5, ChatGPT 4 and Gemini Pro predict (within confidence intervals). A notable exception is scenario 15, "Great idea to go to Thailand in the monsoon season...". Here, the irony that human responders notice and rate is not recognised by ChatGPT 3.5,  ChatGPT 4 and LLaMA2, which wrongly classify the associated sentiment, while Gemini rates it within CI. Instead, scenario 6 ("Just a great idea, going to eat where the neighborhood bullies regularly hang out...") is decently coped by ChatGPTs and Gemini, likely spotting sarcasm leveraging on the "just" keyword (as output by Gemini Pro log). 

Finally, ChatGPT and LLaMA are more aligned with human neutral-to-negative sentiment on scenario 17 ("Paying to visit a place of worship, how absurd"), while Gemini rates it excessively negative. Similarly, Gemini is more negative on Scenario 7 ("I could not appreciate the whole dinner because of a work emergency, but the appetizer was great") and 18 ("Whitewater rafting is an adrenaline experience, but it is not for everyone"). In general, Gemini is relatively more negative than humans in evaluating text sentiment. This negativity may partially explain its greater capability in coping with irony. \\

To better interpret the comparison with LLM results, we also computed the normalized mean rate $R_{\ell}$ assigned by the surveyed individuals for each language, as shown in \cref{fig:human}. The number of questionnaire responses does not allow a thorough quantitative evaluation of biases associated to each language (\textit{cf.} \cref{fig:proportion}). However, the analysis provides preliminary insights in revealing comparable or even greater variability than observed with the LLM models. Notably, native speakers of Polish, Portuguese, and Chinese provided more positive feedback compared to the average, while Russians and English native speakers rate below the average. We do not observe a strong "pessimism" associated to French or Italian as suggested by ChatGPT 3.5, nor particular "optimism" associated to Russian, like ChatGPT 4 would suggest (\textit{cf.} \cref{fig:normalized}a and \cref{fig:normalized}b, respectively). On the other hand, the variability does not overlap with that of Gemini, except for Russian (skewed towards negative) and Chinese, skewed towards positive sentiment, but not as much as Gemini Pro's.

\subsection{Analysis of Gemini Pro censored outputs}
\label{sub:gemini_censorship}

\begin{figure*}%
    \centering
    \subfloat[]{\includegraphics[scale=0.55]{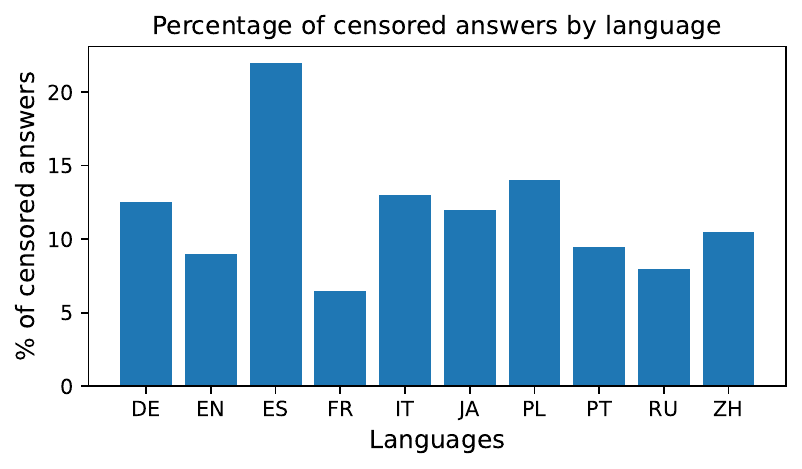}}%
    \qquad
    \subfloat[]{\includegraphics[scale=0.55]{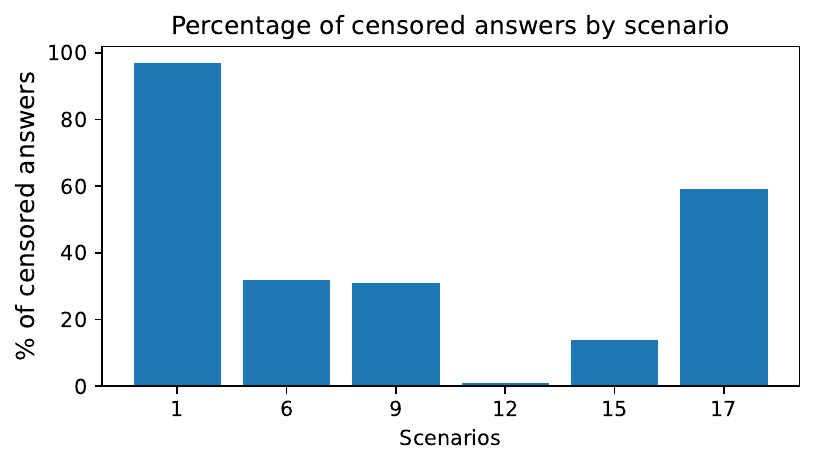}}%
    \caption{Gemini Pro: Proportion of censored answers by language (a) and by scenario (b).}%
    \label{fig:censorship}%
\end{figure*}

As anticipated, Gemini possesses safety filters against undesired content. However, it does not consistently employ them for identical language and scenarios (\cref{fig:censorship}a). For instance, for Scenario 6, the model consistently applies the safety filter in all 10 Japanese runs, but it never applies the filters in the equivalent scenario in Chinese. For all other languages, the model applies the filter in some runs and not in others. Overall, French is the less censored language, Spanish the most censored one, others are filtered in varying proportions but without clear patterns. 

Filters are also triggered on different scenarios with varying strength. For instance, Scenario 1 ("At the restaurant the food was good but the waiter shot someone with a gun") was almost always censored due to the reference to "guns" (as prompted by the model). Other scenarios, depending on different languages, were also often censored, with associated keywords ranging from "bullies" to "worship places", see \cref{fig:censorship}b; however, there is little indication on to how to predict the censorship proportion. 

It is acknowledged and anticipated that \acp{LLM} inherently exhibit a degree of variability, which enables diverse responses to the same prompt. However, it is concerning that even Gemini's assessment of safety appears to be influenced by randomness. Understanding the reasons for such discrepancies would require investigating LLMs kernels and elucidating their decision-making processes, which is currently not possible. We suggest future studies to improve the interpretability of such LLMs, in order to better make sense of censorship behaviors.

\section{Discussion}
\label{sec:discussion}
This paper addresses the problem of automated sentiment analysis performed by modern LLMs, that can be easily employed both for research and for industrial applications. In particular, the paper evaluates their capabilities to navigate vague and ambiguous scenarios, the consistency across models and versions, as well as the qualitative consistency with human responses. We observed that ChatGPT and Gemini usually rate the proposed scenarios in similar fashion to humans, with exceptions related to irony and sarcasm -- features that are notably hard to deal with \cite{farias2017irony}. However, noteworthy differences exist when evaluating various human languages and different ChatGPT versions (3.5 or 4). Moreover, Gemini shows significant rating difference across languages, as well as unexplained censorship behaviour, which may skew its outputs. Instead, LLaMA2 repeatedly scores all scenarios as positive, even in cases that are rather clearly negative (\textit{e.g.,} "Restaurant recommended by a friend. I won't listen to him again", scored 7.8 by LLaMA2). This fact calls for deeper investigation of "positive biases" that may be present in LLMs. We recall that all experiments were performed using default settings, to reproduce common innovation pipelines building on available LLM. 

Tracing back the reason for such discrepancies is extremely important and is demanded to future investigation. This would provide valuable insights on to whether some hidden meaning can be extracted from different languages, or training was performed differently (maybe with some metaparameters tuned differently for each language), or biases were present in the training sets, or whether psychological lexicon \cite{semeraro2023emoatlas} adjustments were not fully integrated in the current LLMs. Addressing these issues would contribute significantly to the field of LLMs, in terms of algorithm development and of ambiguity assessment. Moreover, it would also provide a step forward in the interpretability, transferability and applicability of such models into daily practice, which requires consistency and reliability across languages and versions. By means of questionnaires, we have paved a methodological way to address these issues. Future studies, specifically focused on sociological investigation, may further improve the analysis. \\

We acknowledge that this work is limited in the number of examples provided. Our scenarios were carefully crafted to represent realistic cases that LLMs may process, as well as to contain a degree of ambiguity that can be systematically tested. Extending the examples with scenarios crafted with linguists or extracted from user-generated contents (see, \textit{e.g.}, \cite{ling2016empirical,farha2020arabic}) may provide additional insights. 

Furthermore, we acknowledge that the evaluation of human-scored questionnaires may suffer from hidden biases, e.g. due to the authors' own filter bubble \cite{bruns2019filter}. This note does not debase the quantitative insights obtained by repeated experiments on the LLMs alone, nor significantly impact the qualitative assessment of LLM responses against humans'. However, future studies may improve the statistics about LLM-human discrepancies on ambiguous scenarios and per language, by increasing the cohort of human respondents and by disseminating the surveys to diverse populations. \\

\section{Conclusion}
\label{sec:conclusion}

State-of-the-art \acp{LLM} represent a cutting-edge advancement in \ac{NLP}, offering the promise of effectively navigating the challenges posed by contextual ambiguity, sarcasm, and irony. These models boast the ability to comprehend and interpret nuanced linguistic expressions, enabling them to discern the subtle layers of meaning often inherent in these forms of communication.

In our study, we conducted a comprehensive evaluation of four leading language models, namely ChatGPT 3.5, ChatGPT 4, Gemini Pro and LLaMA2 7b. 
Our assessment involved challenging these models with the task of rating 20 texts that describing a diverse array of scenarios. These scenarios encapsulated a broad spectrum of emotions, ranging from joy and contentment to anger and frustration, providing a nuanced examination of the models' abilities across the emotional landscape.
Our study extended across 10 languages, creating a rich linguistic tapestry that allowed us to gauge the models' consistency and performance in varied linguistic contexts. This multifaceted approach aimed to provide a thorough understanding of the models' capabilities and limitations in handling a wide range of scenarios and emotional nuances across different languages.

The results show that ChatGPT 3.5, ChatGPT 4 and Gemini Pro provide valid prediction of sentiment in most ambiguous scenarios, although they often fail to recognise nuances like irony or sarcasm. However, they still present relevant linguistic biases in associating sentiment to language families. Moreover, the problem of censorship surfaced in Gemini, prompting additional investigation by researchers and regulators alike. On the other hand, we discovered optimistic biases in LLaMA2 predictions, which are consistent across languages but fail to predict negative sentiment associated, e.g, to sarcasm, double meaning of texts, or contrasting occurrences within single scenarios. 

Our study demonstrates that, despite the recent advancements, significant gaps still exist in the applicability, trasferability and interpretability of such algorithms and of their outputs, particularly when vague or ambiguous scenarios are involved. 
We provide a methodology to perform post-hoc sanity checks on LLMs capabilities, to inform programmers about biases and mis-processing, as well as to warn practitioners about the reliability of such outputs to real-world instances. 
We therefore hope to call for a careful review of datasets, methods and application, to minimise linguistic biases, inconsistent performances across models and versions, and obscure safety biases, which would skew the results of automated sentiment analysis in undesired directions. 

For software applications like marketing search or chatbot development, our results can thus provide an additional feature to consider when choosing the kernel LLM, on top of computational cost, efficiency, accuracy in other tasks, and more. Our results can also support informed development and management of automated filters in the context of content moderation \cite{thiago2021fighting} and hate speech detection \cite{sap2019risk}, as well as for industrial applications for companion technologies tailored to different countries.

\printbibliography

@article{wei2022emergent,
  title={Emergent abilities of large language models},
  author={Wei, Jason and Tay, Yi and Bommasani, Rishi and Raffel, Colin and Zoph, Barret and Borgeaud, Sebastian and Yogatama, Dani and Bosma, Maarten and Zhou, Denny and Metzler, Donald and others},
  journal={arXiv:2206.07682},
  year={2022}
}

@article{brown2020language,
  title={Language models are few-shot learners},
  author={Brown, Tom and Mann, Benjamin and Ryder, Nick and Subbiah, Melanie and Kaplan, Jared D and Dhariwal, Prafulla and Neelakantan, Arvind and Shyam, Pranav and Sastry, Girish and Askell, Amanda and others},
  journal={Adv Neur In Proc Sys},
  volume={33},
  pages={1877--1901},
  year={2020}
}

@inproceedings{ling2016empirical,
  title={An empirical, quantitative analysis of the differences between sarcasm and irony},
  author={Ling, Jennifer and Klinger, Roman},
  booktitle={The Semantic Web: ESWC 2016 Satellite Events, Heraklion, Crete, Greece},
  pages={203--216},
  year={2016},
  organization={Springer}
}

@inproceedings{patel2020introduction,
  title={Introduction to common crawl datasets},
  author={Patel, Jay M and Patel, Jay M},
  booktitle={Getting Structured Data from the Internet},
  pages={277--324},
  year={2020},
  publisher={Springer}
}

@inproceedings{farha2020arabic,
  title={From arabic sentiment analysis to sarcasm detection: The arsarcasm dataset},
  author={Farha, Ibrahim Abu and Magdy, Walid},
  booktitle={Proc 4th Workshop Open-Source Arabic Corpora and Processing Tools},
  pages={32--39},
  year={2020}
}

@article{semeraro2023emoatlas,
  title={EmoAtlas: An emotional profiling tool merging psychological lexicons, artificial intelligence and network science},
  author={Semeraro, Alfonso and Vilella, Salvatore and Mohammad, Saif and Ruffo, Giancarlo and Stella, Massimo},
  journal={ResearchSquare preprint},
  year={2023}
}

@article{chakriswaran2019emotion,
  title={Emotion AI-driven sentiment analysis: A survey, future research directions, and open issues},
  author={Chakriswaran, Priya and Vincent, Durai Raj and Srinivasan, Kathiravan and Sharma, Vishal and Chang, Chuan-Yu and Reina, Daniel Guti{\'e}rrez},
  journal={Appl Sci},
  volume={9},
  number={24},
  pages={5462},
  year={2019},
  publisher={MDPI}
}

@inproceedings{mouthami2013sentiment,
  title={Sentiment analysis and classification based on textual reviews},
  author={Mouthami, K and Devi, K Nirmala and Bhaskaran, V Murali},
  booktitle={Int Conf Inf Comm Embedded Sys (ICICES)},
  pages={271--276},
  year={2013},
  organization={IEEE}
}

@article{pang2008opinion,
  title={Opinion mining and sentiment analysis},
  author={Pang, Bo and Lee, Lillian and others},
  journal={Found Trends Inf Ret},
  volume={2},
  number={1--2},
  pages={1--135},
  year={2008},
  publisher={Now Publishers, Inc.}
}

@book{bruns2019filter,
  title={Are filter bubbles real?},
  author={Bruns, Axel},
  year={2019},
  publisher={John Wiley \& Sons}
}

@article{ainslie2023gqa,
  title={GQA: Training Generalized Multi-Query Transformer Models from Multi-Head Checkpoints},
  author={Ainslie, Joshua and Lee-Thorp, James and de Jong, Michiel and Zemlyanskiy, Yury and Lebr{\'o}n, Federico and Sanghai, Sumit},
  journal={arXiv:2305.13245},
  year={2023}
}

@article{eger2019time,
  title={Is it time to swish? Comparing deep learning activation functions across NLP tasks},
  author={Eger, Steffen and Youssef, Paul and Gurevych, Iryna},
  journal={arXiv:1901.02671},
  year={2019}
}

@article{george2023review,
  title={A review of ChatGPT AI's impact on several business sectors},
  author={George, A Shaji and George, AS Hovan},
  journal={PU Int Inno J},
  volume={1},
  number={1},
  pages={9--23},
  year={2023}
}

@article{mohammad2013crowdsourcing,
  title={Crowdsourcing a word--emotion association lexicon},
  author={Mohammad, Saif M and Turney, Peter D},
  journal={Comput Int},
  volume={29},
  number={3},
  pages={436--465},
  year={2013},
  publisher={Wiley Online Library}
}

@inproceedings{maas2011learning,
  title={Learning word vectors for sentiment analysis},
  author={Maas, Andrew and Daly, Raymond E and Pham, Peter T and Huang, Dan and Ng, Andrew Y and Potts, Christopher},
  booktitle={Proc 49th Annu Meeting Ass Comput Ling},
  pages={142--150},
  year={2011}
}

@incollection{mohammad2016sentiment,
  title={Sentiment analysis: Detecting valence, emotions, and other affectual states from text},
  author={Mohammad, Saif M},
  booktitle={Emotion measurement},
  pages={201--237},
  year={2016},
  publisher={Elsevier}
}

@incollection{farias2017irony,
  title={Irony, sarcasm, and sentiment analysis},
  author={Farias, DI Hern{\'a}ndez and Rosso, Paolo},
  booktitle={Sentiment Analysis in Social Networks},
  pages={113--128},
  year={2017},
  publisher={Elsevier}
}

@article{thiago2021fighting,
  title={Fighting hate speech, silencing drag queens? artificial intelligence in content moderation and risks to lgbtq voices online},
  author={Thiago, Dias Oliva and Marcelo, Antonialli Dennys and Gomes, Alessandra},
  journal={Sexuality \& culture},
  volume={25},
  number={2},
  pages={700--732},
  year={2021},
  publisher={Springer Nature BV}
}

@inproceedings{sap2019risk,
  title={The risk of racial bias in hate speech detection},
  author={Sap, Maarten and Card, Dallas and Gabriel, Saadia and Choi, Yejin and Smith, Noah A},
  booktitle={Proc 57th Annu Meeting Ass Comput Ling},
  pages={1668--1678},
  year={2019}
}

@article{team2023gemini,
  title={Gemini: a family of highly capable multimodal models},
  author={Team, Gemini and Anil, Rohan and Borgeaud, Sebastian and Wu, Yonghui and Alayrac, Jean-Baptiste and Yu, Jiahui and Soricut, Radu and Schalkwyk, Johan and Dai, Andrew M and Hauth, Anja and others},
  journal={arXiv:2312.11805},
  year={2023}
}

@article{tang2015deep,
  title={Deep learning for sentiment analysis: successful approaches and future challenges},
  author={Tang, Duyu and Qin, Bing and Liu, Ting},
  journal={Wiley Interd Rev: Data Mining Knowledge Disc},
  volume={5},
  number={6},
  pages={292--303},
  year={2015},
  publisher={Wiley Online Library}
}

@inproceedings{hutto2014vader,
  title={Vader: A parsimonious rule-based model for sentiment analysis of social media text},
  author={Hutto, Clayton and Gilbert, Eric},
  booktitle={Proc Int AAAI Conf Web Social Media},
  volume={8},
  number={1},
  pages={216--225},
  year={2014}
}

@incollection{liu2012survey,
  title={A survey of opinion mining and sentiment analysis},
  author={Liu, Bing and Zhang, Lei},
  booktitle={Mining text data},
  pages={415--463},
  year={2012},
  publisher={Springer}
}

@inproceedings{cambria2013introduction,
  title={An introduction to concept-level sentiment analysis},
  author={Cambria, Erik},
  booktitle={Advances in Soft Computing and Its Applications: 12th Mexican Int Conf Artif Int, MICAI 2013, Mexico City, Mexico},
  pages={478--483},
  year={2013},
  organization={Springer}
}

@article{gebru2022review,
  title={A review on human--machine trust evaluation: Human-centric and machine-centric perspectives},
  author={Gebru, Biniam and Zeleke, Lydia and Blankson, Daniel and Nabil, Mahmoud and Nateghi, Shamila and Homaifar, Abdollah and Tunstel, Edward},
  journal={IEEE T Human-Machine Sys},
  volume={52},
  number={5},
  pages={952--962},
  year={2022},
  publisher={IEEE}
}

@article{piantadosi2012communicative,
  title={The communicative function of ambiguity in language},
  author={Piantadosi, Steven T and Tily, Harry and Gibson, Edward},
  journal={Cognition},
  volume={122},
  number={3},
  pages={280--291},
  year={2012},
  publisher={Elsevier}
}

@article{zhang2023revisiting,
  title={Revisiting Sentiment Analysis for Software Engineering in the Era of Large Language Models},
  author={Zhang, Ting and Irsan, Ivana Clairine and Thung, Ferdian and Lo, David},
  journal={arXiv:2310.11113},
  year={2023}
}

@article{wankhade2022survey,
  title={A survey on sentiment analysis methods, applications, and challenges},
  author={Wankhade, Mayur and Rao, Annavarapu Chandra Sekhara and Kulkarni, Chaitanya},
  journal={Artif Int Rev},
  volume={55},
  number={7},
  pages={5731--5780},
  year={2022},
  publisher={Springer}
}

@article{devlin2018bert,
  title={Bert: Pre-training of deep bidirectional transformers for language understanding},
  author={Devlin, Jacob and Chang, Ming-Wei and Lee, Kenton and Toutanova, Kristina},
  journal={arXiv:1810.04805},
  year={2018}
}

@article{christiano2017deep,
  title={Deep reinforcement learning from human preferences},
  author={Christiano, Paul F and Leike, Jan and Brown, Tom and Martic, Miljan and Legg, Shane and Amodei, Dario},
  journal={Adv Neur In Proc Sys},
  volume={30},
  year={2017}
}

@article{vaswani2017attention,
  title={Attention is all you need},
  author={Vaswani, Ashish and Shazeer, Noam and Parmar, Niki and Uszkoreit, Jakob and Jones, Llion and Gomez, Aidan N and Kaiser, {\L}ukasz and Polosukhin, Illia},
  journal={Adv Neur In Proc Sys},
  volume={30},
  year={2017}
}

@article{liang2022holistic,
  title={Holistic evaluation of language models},
  author={Liang, Percy and Bommasani, Rishi and Lee, Tony and Tsipras, Dimitris and Soylu, Dilara and Yasunaga, Michihiro and Zhang, Yian and Narayanan, Deepak and Wu, Yuhuai and Kumar, Ananya and others},
  journal={arXiv:2211.09110},
  year={2022}
}

@article{touvron2023llama,
  title={Llama 2: Open Foundation and Fine-Tuned Chat Models},
  author={Touvron, Hugo and Martin, Louis and Stone, Kevin and Albert, Peter and Almahairi, Amjad and Babaei, Yasmine and Bashlykov, Nikolay and Batra, Soumya and Bhargava, Prajjwal and Bhosale, Shruti and others},
  journal={arXiv:2307.09288},
  year={2023}
}

@inproceedings{gillioz2020overview,
  title={Overview of the Transformer-based Models for NLP Tasks},
  author={Gillioz, Anthony and Casas, Jacky and Mugellini, Elena and Abou Khaled, Omar},
  booktitle={15th Conference FedCSIS},
  pages={179--183},
  year={2020},
  organization={IEEE}
}

@online{huggingFace,
    author = "HuggingFace",
    title = "Llama 2",
    url  = "https://huggingface.co/meta-llama/Llama-2-7b",
    year = "2023"
}

@article{cambria2017sentiment,
  title={Sentiment analysis is a big suitcase},
  author={Cambria, Erik and Poria, Soujanya and Gelbukh, Alexander and Thelwall, Mike},
  journal={IEEE Intell Sys},
  volume={32},
  number={6},
  pages={74--80},
  year={2017},
  publisher={IEEE}
}

@online{gemini,
    author = "Google",
    title = "Introducing Gemini: our largest and most capable AI model",
    url  = "https://blog.google/technology/ai/google-gemini-ai/",
    year = "2023"
}

@online{chatgpt,
    author = "OpenAI",
    title = "Introducing ChatGPT",
    url  = "https://openai.com/blog/chatgpt",
    year = "2023"
}

@online{semeval,
    author = "SemEval",
    title = "International Workshop on Semantic Evaluation",
    url  = "https://semeval.github.io/",
    year = "2024"
}

\end{document}